\documentclass[letterpaper,10pt]{article} 
\usepackage{graphicx}
\usepackage{subfigure}
\usepackage{float}
\floatstyle{plain} 
\newfloat{Output}{H}{myc}

\usepackage[cmex10]{amsmath}
\usepackage[normalem]{ulem}
\usepackage[utf8]{inputenc}                              
\newcommand{\sig}{\text{sig}}

\begin{document}
\title{A Tutorial on Deep Neural Networks for Intelligent Systems}

\author{Juan C. Cuevas-Tello$^{1,2}$ \and Manuel Valenzuela-Rend\'on $^1$ \and Juan A. Nolazco-Flores$^1$ \\
\small $^1$Tecnol\'ogico de Monterrey, Campus Monterrey \\ 
\small Av.~Eugenio Garza Sada 2501 Sur,  C.P.~64849, \\ 
\small Monterrey, N.L., Mexico \\
\small \{cuevastello,valenzuela,jnolazco\}@itesm.mx \\
\small $^2$Engineering Faculty, UASLP \\ 
\small Dr. Manuel Nava No. 8, Zona Universitaria, C.P.~78290  \\ 
\small San Luis Potosi, SLP, Mexico \\ 
\small cuevas@uaslp.mx \\
}


\maketitle

\begin{abstract}
Developing Intelligent Systems involves artificial intelligence approaches including artificial neural networks. Here, we present a tutorial of Deep Neural Networks (DNNs), and some insights about the origin of the term ``deep''; references to deep learning are also given. Restricted Boltzmann Machines, which are the core of DNNs, are discussed in detail. An example of a simple two-layer network, performing unsupervised learning for unlabeled data, is shown. Deep Belief Networks (DBNs),  which are used to build networks with more than two layers, are also described. Moreover, examples for supervised learning with DNNs performing simple prediction and classification tasks, are presented and explained. This tutorial includes two intelligent pattern recognition applications: handwritten digits (benchmark known as MNIST) and speech recognition.
\end{abstract}

\section{Introduction}\label{intro}
Intelligent systems involve artificial intelligence approaches including artificial neural networks. This paper focus mainly on Deep Neural Networks (DNNs). 

The core of DNNs are the Restricted Boltzmann Machines (RBMs) proposed by Smolensky \cite{Smolensky:1986:FHT,Hinton:2002:PoE}, and widely studied by Hinton et al.\  \cite{Hinton:2006:Science,Hinton:2006:DBN,Hinton:2010:Guide}, where the term \emph{deep} comes from Deep Beliefs Networks (DBN) \cite{Hinton:2006:DBN}. The next section describes the relationship among RBMs, DBN and DNNs.

Nowadays, the term \emph{Deep Learning} (DL) is becoming popular in the machine learning literature \cite{Le:2012:DeepLearning,Andrew-NG:2013:DeepLearning,Schmidhuber:2015:DL}. However, DL mainly refers to Deep Neural Networks (DNNs) and in particular to DBNs and RBMs \cite{Le:2012:DeepLearning}. Some work related to DL is focusing on high performance computing to speed up the learning of DNNs, i.e.\ Graphics Processing Units (known as GPUs), Message Passing Interface (MPI)  among other parallelization technologies \cite{Andrew-NG:2013:DeepLearning}. 

A wide survey on artificial intelligence and in particular DL has been published recently, which covers DNNs, Convolutional Neural Networks, Recurrent Neural Networks, among many other learning strategies \cite{Schmidhuber:2015:DL}.

A Restricted Boltzmann Machine (RBM) is defined as
\begin{quotation}
   a single layer of hidden units which are not connected to each other and have undirected, symmetrical connections to a layer of visible units. The visible units and the hidden states are sampled from their
conditional distribution using Gibbs sampling by running a Markov chain until it reaches its stationary distribution. The learning rule is the same as the maximum likelihood learning rule [contrastive divergence] for the infinite logistic belief net with tied weights \cite{Hinton:2006:DBN}. 
\end{quotation}
Products of Experts (PoE) and Boltzmann machines are probabilistic generative models, and their intersection comes up with RBMs \cite{Hinton:2002:PoE}. Learning by contrastive divergence of PoE is the basis of the learning algorithm of DBNs \cite{Hinton:2002:PoE,Hinton:2006:DBN}.

We recommend \cite{Fischer:2014:RBM} as a gentle introduction that explains the training of RBMs and their relationship to graphical models including Markov Random Fields (MRFs); it also presents Markov chains to explain how  a RBM draws samples from probability distributions such as Gibbs distribution of a MRF.

The building blocks of a RBM are binary stochastic neurons \cite{Hinton:2006:DBN}. Nevertheless, there are several ways to define real-valued visible neurons, where Gaussian-Binary-RBM are widely used \cite{Fischer:2014:RBM}.

We use a publicly available MATLAB\textsuperscript{\textregistered{}}/Octave toolbox for RBMs developed by Tanaka and Okutomi~\cite{Tanaka:2014:RBM-toolbox-Matlab}. This toolbox implements sparsity \cite{Lee:2008:Sparse-RBM}, dropout \cite{Hinton:2013:Dropout} and a novel inference for RBM \cite{Tanaka:2014:RBM-toolbox-Matlab}.

The main contribution of this tutorial are the DNNs examples along the source code (Matlab/Octave) to build intelligent systems. Therefore, the spirit of this tutorial is that people can easily execute the examples and see what kind of results are obtained. There are examples with either unsupervised or supervised learning, and examples for prediction and classification tasks are also provided. Moreover, the parameter setting of DNNs with an example is shown.

This tutorial is organized as follows: The following section (\S\ref{sec:RBM}) describes RBMs. Section~\S\ref{sec:toolbox} describes the toolbox developed by Tanaka and Okutomi~\cite{Tanaka:2014:RBM-toolbox-Matlab} and the database MNIST. Section~\S\ref{sec:parameter-setting} presents a discussion about the parameter settings of DNNs. Section~\S\ref{sec:examples} explains some simple examples of DNNs. The last section presents speech processing with DNNs; \S\ref{sect:speech}. Finally, a summary and references are given.

\section{Restricted Boltzmann Machines (RBMs)}\label{sec:RBM}
\begin{figure}
   \centering
   \includegraphics[width=5in]{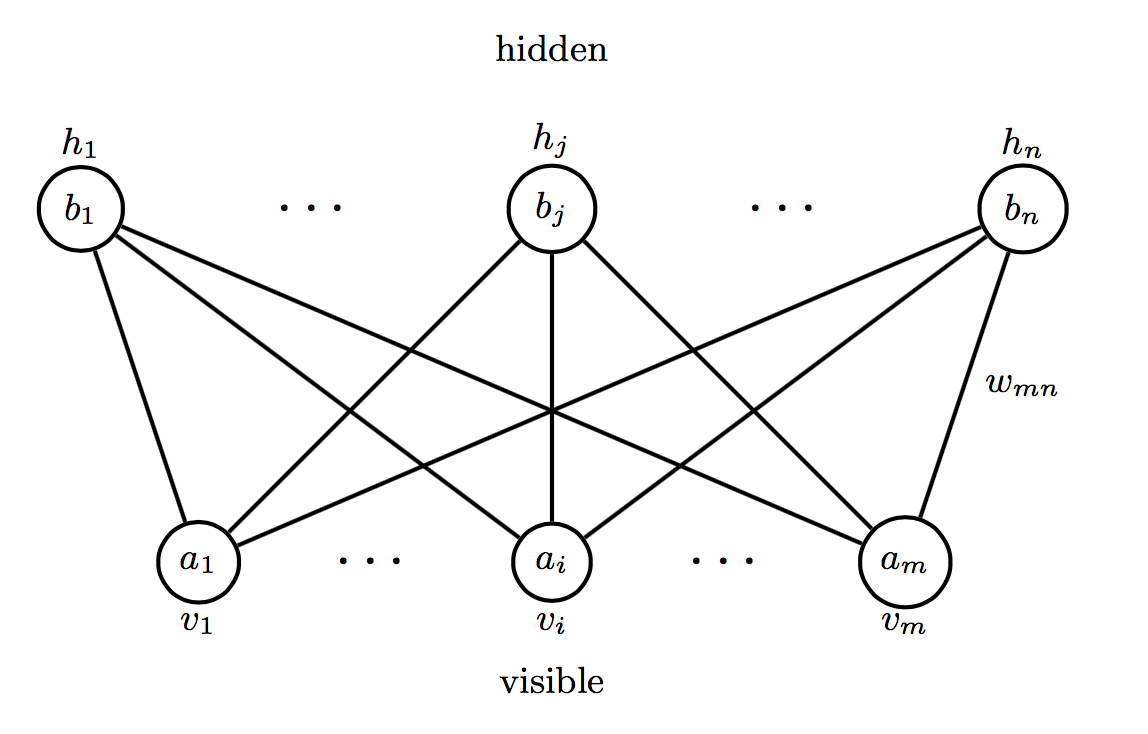}
   \caption{\label{fig:RBM} A RBM with a visible layer and a hidden layer.}
\end{figure}

A RBM is depicted in Fig.~\ref{fig:RBM}. The visible layer is the input, unlabeled data, to the neural network. The hidden layer grabs features from the input data, and each neuron captures a different feature \cite{Hinton:2006:DBN}. By definition, a RBM is a bipartite undirected graph. A RBM has $m$ visible units $\vec{V}=(V_1,V_2,\ldots,V_m)$, the input data, and $n$ hidden units $\vec{H}=(H_1,H_2,\ldots,H_n)$, the features \cite{Fischer:2014:RBM}. A joint configuration, $(\vec{v},\vec{h})$ of the visible and hidden units has an energy given by \cite{Hopfield:1982:NN}
\begin{equation}
\label{eq:energy}
   E(\vec{v},\vec{h}) = 
   - \sum_{i=1}^m a_i v_i 
   - \sum_{j=1}^n b_j h_j
   - \sum_{i=1}^m\sum_{j=1}^n v_i h_j w_{ij}\ ,
\end{equation}
where $v_i$ and $h_j$ are the binary states of the visible and hidden units, respectively; $a_i$, $b_j$ are the biases, and $w_{ij}$ is a real valued weight associated with each edge in the network \cite{Hinton:2010:Guide}, see Fig.~\ref{fig:RBM}.

\begin{figure}
\centering
\includegraphics[width=4in]{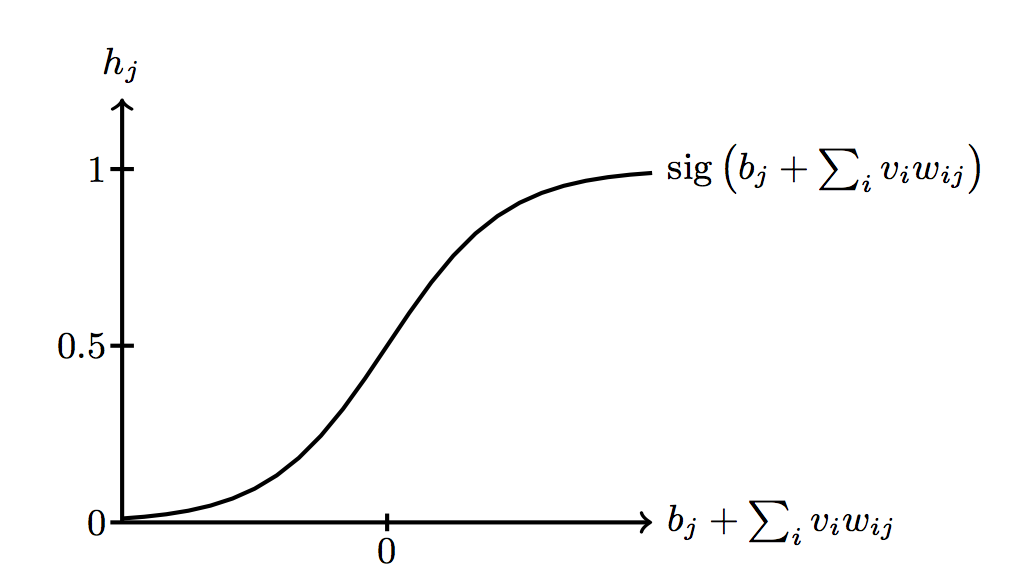}
\caption{\label{fig:neuron} The building block of a RBM: a binary stochastic neuron. In this example, visible to hidden, $h_j$ is the probability of producing a spike \cite{Hinton:2010:Guide}.}
\end{figure}

The building block of a RBM is a binary stochastic neuron \cite{Hinton:2006:DBN}. Fig.~\ref{fig:neuron} shows how to obtain the state of a hidden neuron given a visible layer (data).

A RBM can be seen as a stochastic neural network. First, weights $w_{ij}$ are randomly initialized. Then, the data to be learned is set at the visible layer; this data can be an image, a signal, etcetera. Now, the state of the neurons at the hidden layer is obtained by
\begin{equation}
\label{eq:v2h}
    p(h_j=1|\vec{v})=\sig\left(b_j+\sum_i v_i w_{ij}\right),
\end{equation}
so the conditional probability of $h_j$ being $1$ is the firing rate of a stochastic neuron with a sigmoid activation function, 
$\sig(x)=1/(1-e^{-x})$, 
see Fig.~\ref{fig:neuron}. This step, visible to hidden, is represented as $\langle v_i h_j \rangle^0$, at time $t=0$ \cite{Hinton:2006:DBN,Fischer:2014:RBM,Tanaka:2014:RBM-toolbox-Matlab}.

\subsection{Contrastive Divergence algorithm}
\begin{figure}
\centering
\includegraphics[width=5in]{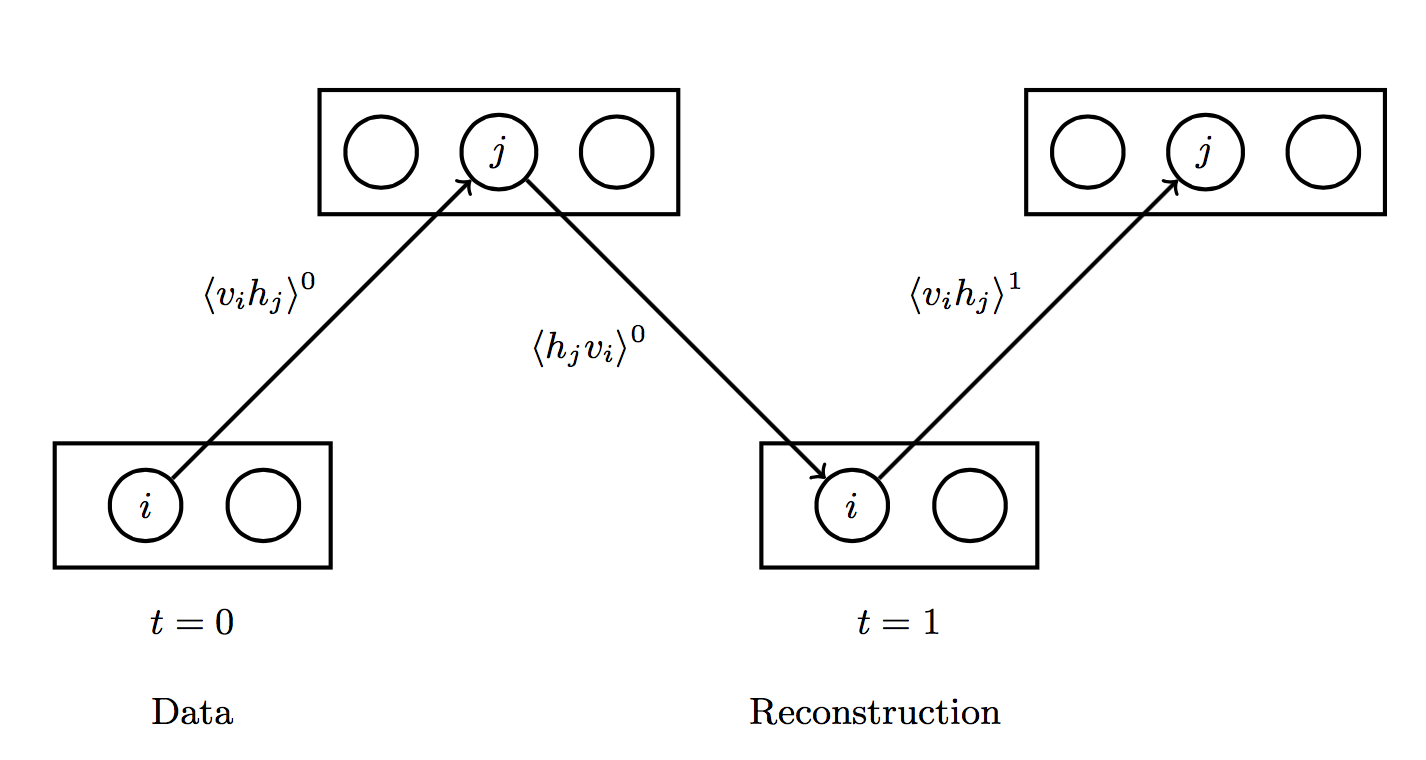}
\caption{\label{fig:CD} Contrastive Divergence (CD) algorithm.}
\end{figure}

Learning in a RBM is achieved by the Contrastive Divergence (CD) algorithm, see Fig.~\ref{fig:CD}~\cite{Hinton:2006:DBN}. The first step of the CD algorithm is $\langle v_i h_j\rangle^0$, as shown above. The next step is the ``reconstruction'' of the visible layer by
\begin{equation}
\label{eq:h2v}
    p(v_i=1|\vec{h})=\sig\left(a_i+\sum_{j}h_j w_{i,j}\right),
\end{equation}
i.e., hidden to visible. This step is denoted as $\langle h_j v_i \rangle ^0$. The new state of the hidden layer is obtained using the result of the reconstruction as the input data, and this step is denoted as $\langle v_i h_j \rangle ^1$; at time $t=1$. Finally, the weights and biases are adjusted in the following way \cite{Hinton:2006:DBN}:

\begin{align}
    \Delta w_{ij} &= 
    \varepsilon\left(
    	\langle v_i h_j\rangle^0 - \langle v_i h_j\rangle^1 
        \right) 
    \label{eq:weightsW}; \\
    \Delta a_i &= \varepsilon\left(v_i^0-v_i^1\right) 
    \label{eq:weightsa}; \\
    \Delta b_j &= \varepsilon\left(h_j^0-h_j^1\right) 
    \label{eq:weightsb};
\end{align}
where $\varepsilon$ is the learning rate. RBMs find better models if more steps of the CD algorithm are performed; $\text{CD}_k$ is used to denote the learning in $k$ steps/iterations \cite{Hinton:2010:Guide}. The CD algorithm is summarized in Algorithm~1, and it uses the complete training data, \textit{batch learning} \cite{Fischer:2014:RBM}.

\begin{figure}  
\includegraphics[width=5in]{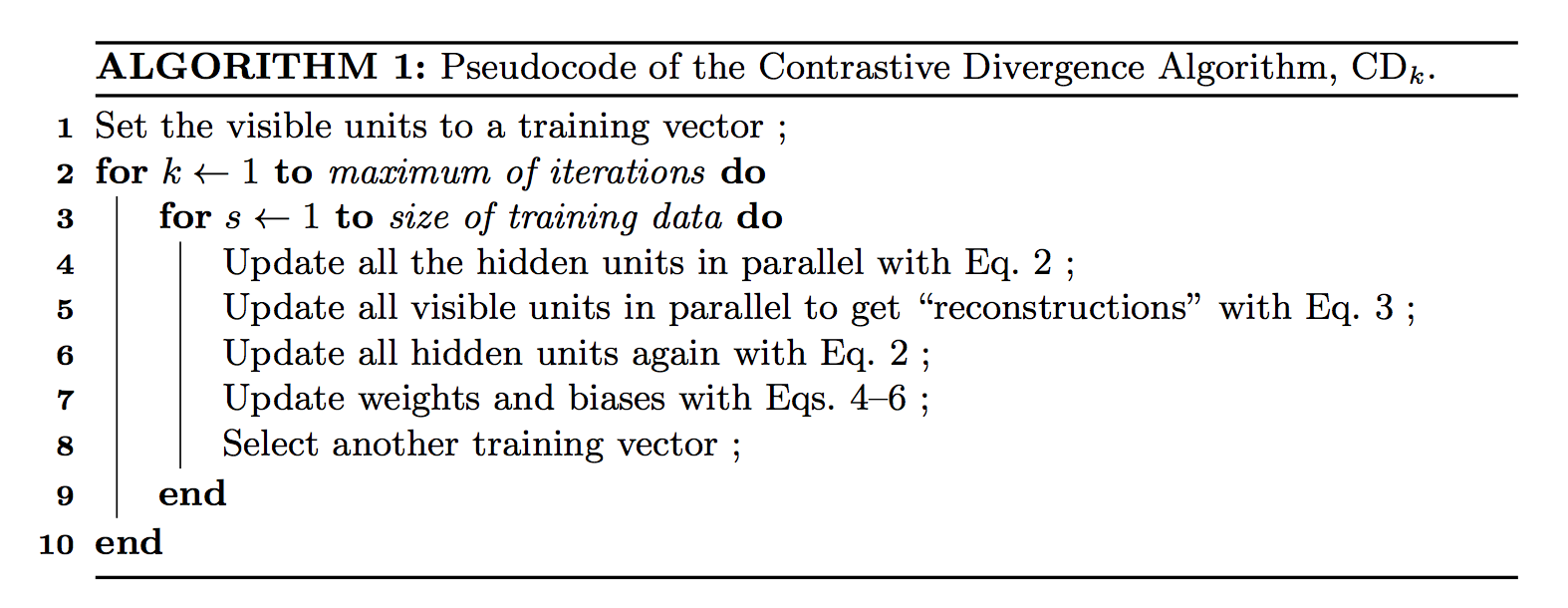}
\end{figure}

\subsection{Deep Belief Network}
\begin{figure}
\centering
\includegraphics[width=3in]{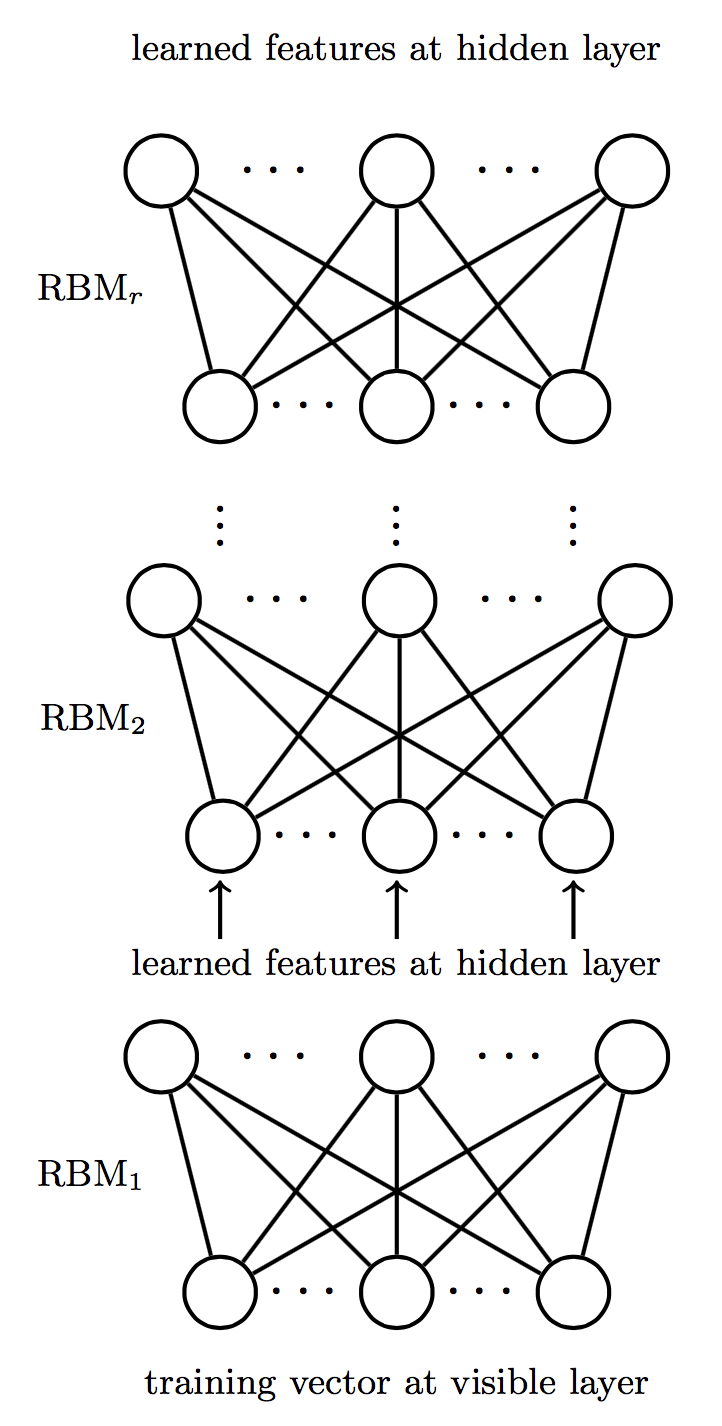}
\caption{\label{fig:DBN} An infinite belief network. The more RBMs, the deeper the learning; i.e.\ Deep Belief Networks (DBN).}
\end{figure}

\begin{figure}
\centering
\includegraphics[width=3in]{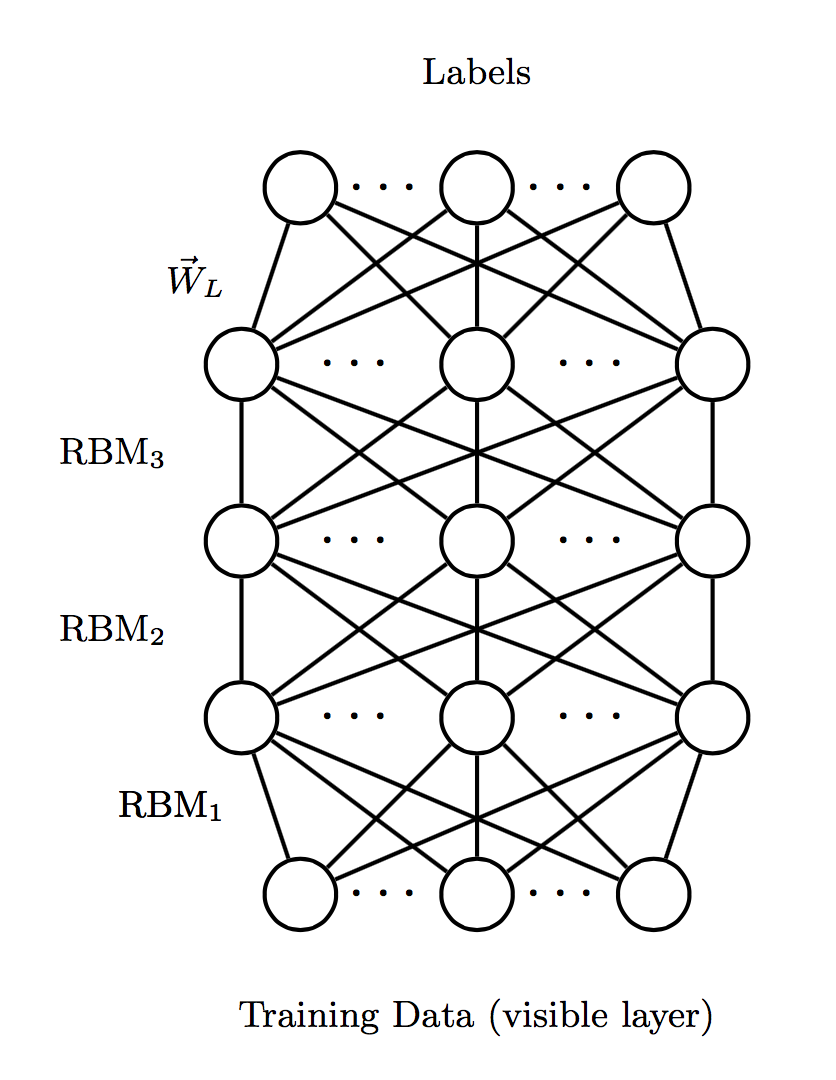}
\caption{\label{fig:DBN2} An hybrid DBN for supervised learning.}
\end{figure}

A Deep Belief Network (DBN) \cite{Hinton:2006:DBN} is depicted in Fig.~\ref{fig:DBN}. Comparing Fig.~\ref{fig:RBM} with Fig.~\ref{fig:DBN}, we can see that a DBN is built by stacking RBMs. Thus, the more levels the DBN has, the deeper the DBN is. The hidden neurons in a $\text{RBM}_1$ capture the features from the visible neurons. Then, those features become the input to $\text{RBM}_2$, and so on until the $\text{RBM}_r$ is reached; see also Fig.~\ref{fig:DBN2}. A DBN extracts features from features in an unsupervised manner (deep learning).

A hybrid DBN has been proposed for supervised learning, see Fig.~\ref{fig:DBN2}. This network adds labels to the top layer. The weights $\vec{W}_L$ between the top level and the last layer of hidden neurons, associative memory, are learned in a supervised manner. This process is called fine-tunning \cite{Hinton:2006:DBN}, and it can be achieved by many different algorithms including backpropagation \cite{Rumelhart:1986:Backpropagation,Bishop:1995:book,Rojas:1996:book,Haykin:1999:book}. This hybrid DBN is referred as Deep Neural Networks~\cite{Lee:2008:Sparse-RBM,Hinton:2013:Dropout,Fischer:2014:RBM,Tanaka:2014:RBM-toolbox-Matlab}.

\begin{figure}
\centering
\includegraphics[width=4in]{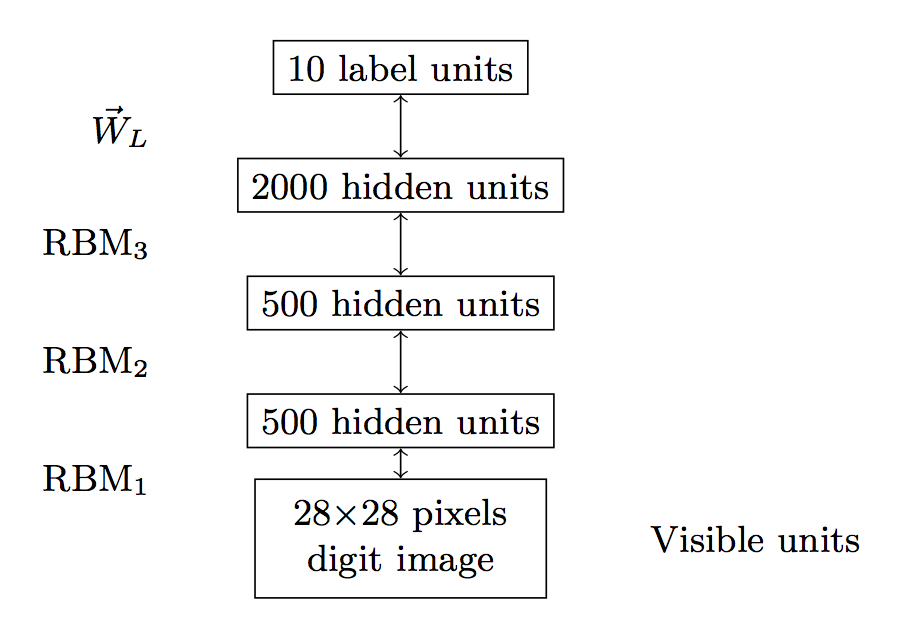}
\caption{\label{fig:DBN3} An hybrid DBN for supervised learning \cite{Hinton:2006:DBN}; the MNIST database.}
\end{figure}

Hinton et al. applied a DNN to the MINST handwritten digits database\footnote{http://yann.lecun.com/exdb/mnist/}~\cite{Hinton:2006:DBN}, see Fig.~\ref{fig:DBN3}. At that time, the DNN produced the best performance with an error rate of 1.25\% compared with other methods including Support Vector Machines (SVM) which had an error rate of 1.4\%~\cite{Hinton:2006:DBN}.

\section{Toolbox for DNN}\label{sec:toolbox}
We use a publicly available toolbox for MATLAB\textsuperscript{\textregistered{}} developed by Tanaka and Okutomi~\cite{Tanaka:2014:RBM-toolbox-Matlab}, and which can be downloaded online\footnote{http://www.mathworks.com/matlabcentral/fileexchange/42853-deep-neural-network}. This toolbox is based on \cite{Hinton:2006:DBN}. This toolbox includes sparsity \cite{Lee:2008:Sparse-RBM}, dropout \cite{Hinton:2013:Dropout} and a novel inference for RBM devised by Tanaka~\cite{Tanaka:2014:RBM-toolbox-Matlab}. Once the toolbox has been downloaded and unzipped, it will generate the following directories:
\begin{itemize}
\item \texttt{/DeepNeuralNetwork/}
\item \texttt{/DeepNeuralNetwork/mnist}
\end{itemize}

\subsection{MNIST}\label{sec:MNIST}
The MNIST database\footnote{\texttt{http://yann.lecun.com/exdb/mnist/}} of handwritten digits has a training set of 60,000 examples, and a test set of 10,000 examples. Once the MNIST has been downloaded and unzipped, we will come up with the following files:
\begin{itemize}
\item \texttt{train-images-idx3-ubyte}: training set images
\item \texttt{train-labels-idx1-ubyte}: training set labels
\item \texttt{t10k-images-idx3-ubyte}:  test set images
\item \texttt{t10k-labels-idx1-ubyte}:  test set labels
\end{itemize}
Note that when you uncompress the \texttt{*.gz} files, then you will need to check the file names, and replace ``.'' by ``\texttt{-}''. You must locate the files within the \texttt{/DeepNeuralNetwork/mnist/} directory.

\subsection{Running the example: DNN-MNIST}
\begin{figure}
\centering
\includegraphics[width=4in]{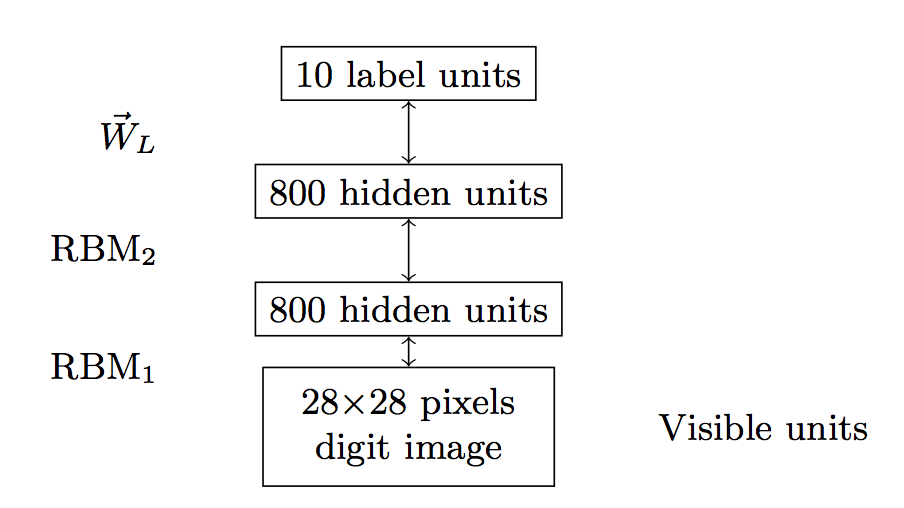}
\caption{\label{fig:DBN-DNN-MNIST} Another DNN architecture for the MNIST database \cite{Tanaka:2014:RBM-toolbox-Matlab}.}
\end{figure}

The file \texttt{/mnist/testMNIST.m} is the main file of the example provided by the toolbox to train a DNN for the MNIST database. The example uses a hybrid network with only two hidden layers of 800 neurons each layer, see Fig.~\ref{fig:DBN-DNN-MNIST}. We have tested the toolbox on Octave\footnote{\texttt{https://www.gnu.org/software/octave/}} 3.2.4 and MATLAB\textsuperscript{\textregistered{}} 7.11.0.584 (2010b), both in Linux Operating Systems.

The script \texttt{testMNIST.m} will generate the file \texttt{mnistbbdbn.mat}  with the DNN already trained. Once \texttt{testMNIST.m} has finished it will appear something like:
\begin{itemize}
\item For training data: rmse = 0.0155251; ErrorRate = 0.00196667  (0.196\%);  Tanaka et al.\ reported 0.158\%~\cite{Tanaka:2014:RBM-toolbox-Matlab}.
\item For test data: rmse = 0.0552593; ErrorRate = 0.0161 (1.6\%);  Tanaka et al.\ reported  1.76\%~\cite{Tanaka:2014:RBM-toolbox-Matlab}.
\end{itemize}

The computational time required to train 60,000 MNIST examples and 10,000 examples for testing is about 3 days on a computer with 384~GB memory, 4~CPUs 2.3GHz with 16~cores each (total of 64~cores).

\subsection{Understanding the toolbox example for MNIST}
\begin{figure}
\centering
\includegraphics[width=5.3in]{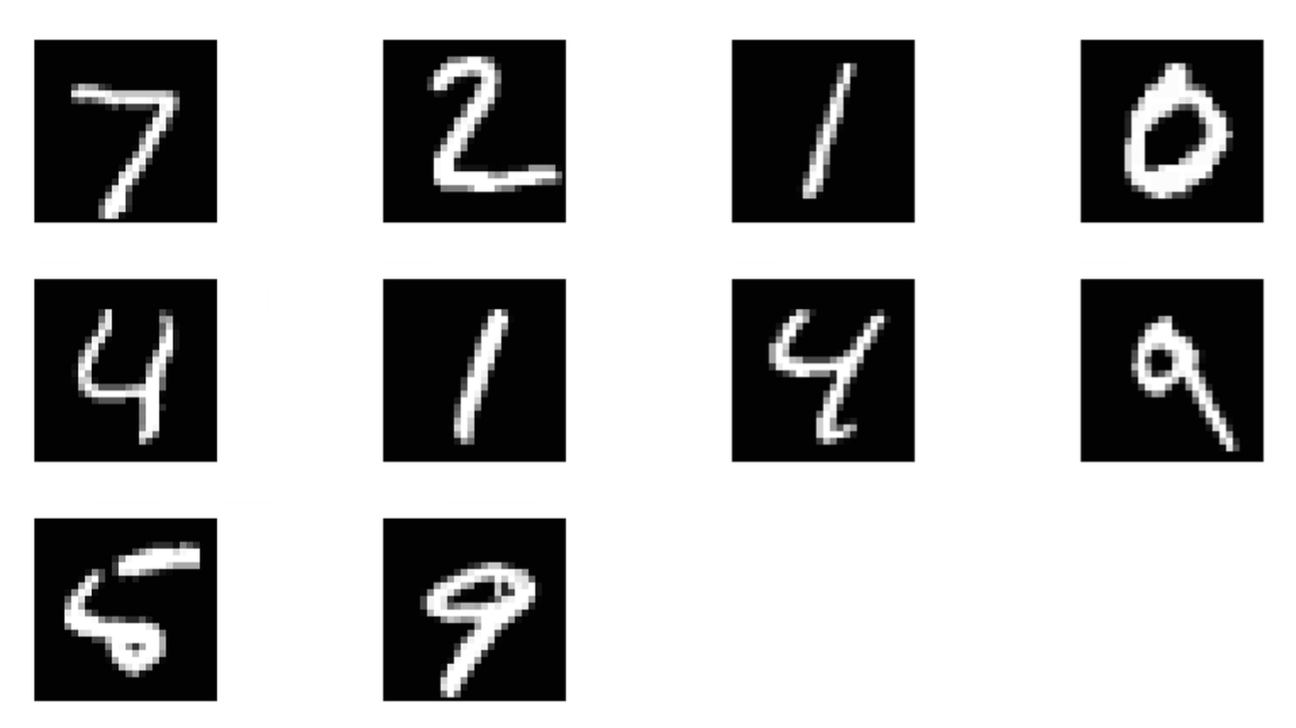}
\caption{\label{fig:MNIST-10-testing-digits} The first 10 testing examples of the MNIST database. The first one is the digit 7, then 2; the last one is the digit 9.}
\end{figure}

Once the example script (\texttt{testMNIST.m}) has been successfully executed, we run the script in Fig.~\ref{fig:test-MINIST}.
\begin{figure}[H]
\centering
 \begin{verbatim}
 % Matlab/Octave script
   mnistfilenames = cell(4,1);
   mnistfilenames{1} = 'train-images-idx3-ubyte'; 
   mnistfilenames{2} = 'train-labels-idx1-ubyte';
   mnistfilenames{3} = 't10k-images-idx3-ubyte';
   mnistfilenames{4} = 't10k-labels-idx1-ubyte';
   [TrainImages TrainLabels TestImages TestLabels]=mnistread(mnistfilenames);
   % load data for training and testing from files
   load mnistbbdbn;    % load the trained DNN 
   dbn = bbdbn; % set dbn as the trained net
   N = 10;  %number of test data to analyze
   IN = TestImages(1:N,:);     %load only N records of testing data
   OUT = TestLabels(1:N,:);  %load only N records of testing data
   for i=1:N,
      imshow(reshape(IN(i,:),28,28));
      name = ['print img-training-',num2str(i),'.jpg -djpeg']
      eval(name); %save the plot, file in jpeg format
   end
   % v2h: get the output of the DNN
   out = v2h( dbn, IN ); 
   % get the maximum values (m) of out and indexes (ind)
   [m ind] = max(out,[],2); 
   out = zeros(size(out)); % initialize the variable out
   % Now, fill with ones where the maximum values where located (ind):
   for i=1:size(out,1)
       out(i,ind(i)) = 1;
   end
   % Now compare out vs OUT. Let say the output of the DNN (out) vs 
   % the desired output (OUT)
   ErrorRate = abs(OUT-out);  % analytically compare OUT vs out
   % sum(ErrorRate,2) performs the sum in two dimensions, 
   % first by row then the resulting column.
   % It is divided by 2; if some output fails, the sum will count twice.
   % mean gives us the percentage of error, known as error rate.
   ErrorRate = mean(sum(ErrorRate,2)/2) % finally the Error rate is obtained
 \end{verbatim}
 \caption{Matlab/Octave script: analysis of $N=10$ test images from MNIST database.}
 \label{fig:test-MINIST}
 \end{figure}
This script generates $N=10$ images via \verb|imshow|, see Fig.~\ref{fig:MNIST-10-testing-digits}. The images are part of the 10 first testing samples.
Each image is stored in a vector of size 784, which corresponds to an image size of 28$\times$28 pixels. And each pixel stores a number between 0 and 255, where 0 means background (white) and 255 means foreground (black); see Fig.~\ref{fig:MNIST-10-testing-digits}.
\begin{figure}[H]
\centering
\includegraphics[width=5in]{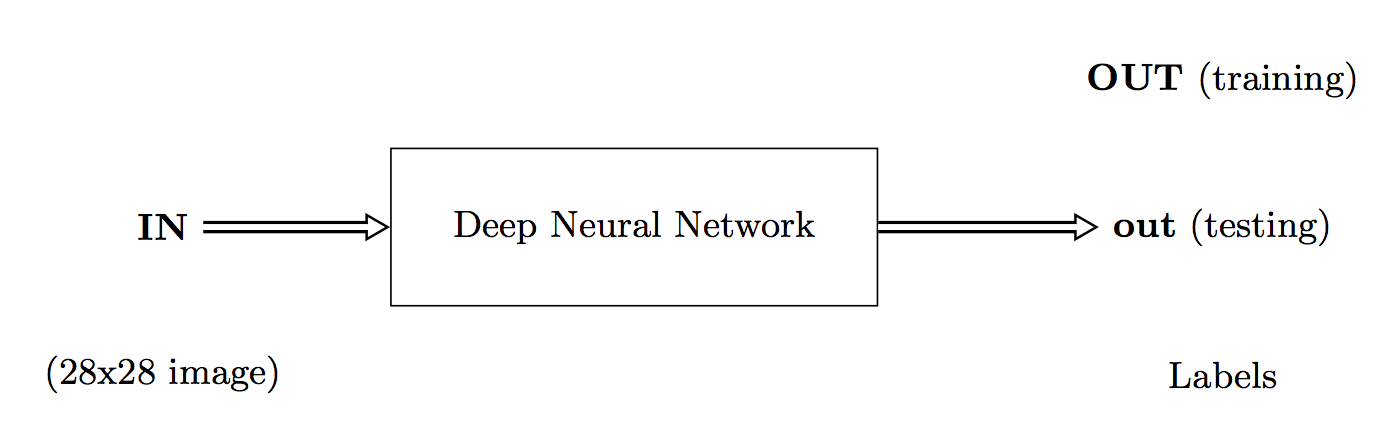}
\caption{\label{fig:DNN-MNIST} Inputs and outputs of a DNN for the MNIST database.}
\end{figure}

In Fig.~\ref{fig:DNN-MNIST}, we depict the inputs and outputs of the DNN for the MNIST database. The variable \textbf{IN} represents the inputs for either training or testing samples. The variable \textbf{OUT} is the output for training, and the variable \textbf{out} the output for testing. Both output variables represent the labels of the digits to learn/recognize (digits from 0 to 9).

For example, the script in Fig. \ref{fig:test-MINIST} generates the following result:
\begin{Output}[H]
\centering
\begin{verbatim}
out =
   		0   0   0   0   0   0   0   1   0   0
   		0   0   1   0   0   0   0   0   0   0
   		0   1   0   0   0   0   0   0   0   0
   		1   0   0   0   0   0   0   0   0   0
   		0   0   0   0   1   0   0   0   0   0
   		0   1   0   0   0   0   0   0   0   0
   		0   0   0   0   1   0   0   0   0   0
   		0   0   0   0   0   0   0   0   0   1
   		0   0   0   0   0   1   0   0   0   0
   		0   0   0   0   0   0   0   0   0   1
\end{verbatim}
\end{Output}
The variable \textbf{out} represents the labels. Each row correspond to each digit image\footnote{Note that there are 60,000 images for training, 10,000 images for testing and 10 images for illustration purposes (Fig.~\ref{fig:MNIST-10-testing-digits}), i.e.\ only 10~rows.}, and each column means the activation or not of the digits 0--9. That is, the first column represents the digit~0, and the last column the digit~9. For example, see top-left in Fig.~\ref{fig:MNIST-10-testing-digits}, the image representing the digit 7 activates only column 8 (i.e.\ 0   0   0   0   0   0   0   1   0   0) of the first row of variable \textbf{out}. The image at the right side of digit 7 corresponds to digit 2, so the third column is activated on the second row of variable \textbf{out}, and so on. In this example the \texttt{ErrorRate} is zero, because the first ten samples of testing are all recognized successfully. Let us create an hypothetical scenario where the DNN fails to recognize the first image (digit~7), i.e.\ imagine that the output looks like the following:
\begin{Output}[H]
\begin{verbatim}
out =
   			0   0   0   0   0   0   1   0   0   0
   			0   0   1   0   0   0   0   0   0   0
   			0   1   0   0   0   0   0   0   0   0
   			1   0   0   0   0   0   0   0   0   0
   			0   0   0   0   1   0   0   0   0   0
   			0   1   0   0   0   0   0   0   0   0
   			0   0   0   0   1   0   0   0   0   0
   			0   0   0   0   0   0   0   0   0   1
   			0   0   0   0   0   1   0   0   0   0
   			0   0   0   0   0   0   0   0   0   1
\end{verbatim}
\end{Output}
Look at the first row, now this row indicates that the first digit image in Fig.~\ref{fig:MNIST-10-testing-digits} corresponds to the digit 6, instead of digit 7. Therefore, \verb|Errorrate| measures this error via the \verb|abs| function along the difference between the desired output \verb|OUT| and the output of the DNN, which is \verb|out|. Then the error rate is obtained as follows:
\begin{Output}[H]
\begin{verbatim}
ErrorRate = abs(OUT-out)
ErrorRate =
   			0   0   0   0   0   0   1   1   0   0
   			0   0   0   0   0   0   0   0   0   0
   			0   0   0   0   0   0   0   0   0   0
   			0   0   0   0   0   0   0   0   0   0
   			0   0   0   0   0   0   0   0   0   0
   			0   0   0   0   0   0   0   0   0   0
   			0   0   0   0   0   0   0   0   0   0
   			0   0   0   0   0   0   0   0   0   0
   			0   0   0   0   0   0   0   0   0   0
   			0   0   0   0   0   0   0   0   0   0
\end{verbatim}
\end{Output}
Now, we sum out by row. This is to detect how many differences are found by test sample.
\begin{Output}[H]
\begin{verbatim}
sum(ErrorRate,2)
ans =
   		2
   		0
   		0
   		0
   		0
   		0
   		0
   		0
   		0
   		0
\end{verbatim}
\end{Output}
If an error exists, then the result is divided by 2. This is because if there is some difference per each sample we will have two 1's as show above.
\begin{Output}[H]
\begin{verbatim}
sum(ErrorRate,2)/2
ans =
   		1
   		0
   		0
   		0
   		0
   		0
   		0
   		0
   		0
   		0 

\end{verbatim}
\end{Output}
Finally, the error rate is given as the mean value:
\begin{Output}[H]
\begin{verbatim}
mean(sum(ErrorRate,2)/2)
ans =  0.10000
\end{verbatim}
\end{Output}

Since the DNN fails to recognize one digit out of ten, the error rate is 10\%  (i.e.\ 0.10000).

\section{Parameter Setting}\label{sec:parameter-setting}
There are several parameters to set up when working with DNNs, including statistics to monitor the contrastive divergence algorithm, batch sizes, monitoring overfitting (iterations), learning rate $\varepsilon$, initial weights, number of hidden unit and hidden layers, types of units (e.g.\ binary or Gaussian), dropout, among others~\cite{Lee:2008:Sparse-RBM,Hinton:2010:Guide,Hinton:2013:Dropout,Fischer:2014:RBM}. In practice, we can only focus on the following:
\begin{itemize}
 \item Maximum of iterations (\verb|MaxIter|), which is also know as $k$ for the contrastive divergence algorithm.
 \item The learning rate $\epsilon$ (\verb|StepRatio|).
 \item Type units (e.g.~Bernoulli or Gaussian distribution).
\end{itemize}
We analyze the impact of these DNN parameters through the XOR example; see \S\ref{sec:examples:XOR}~below.

\section{Further Examples}\label{sec:examples}
Besides the MNIST database example, described above, this section presents examples for unsupervised and supervised learning; including prediction and classification tasks.
\subsection{Unsupervised learning}\label{sec:examples:unsupervised}
\begin{figure}
   \centering
   \includegraphics[width=4.5in]{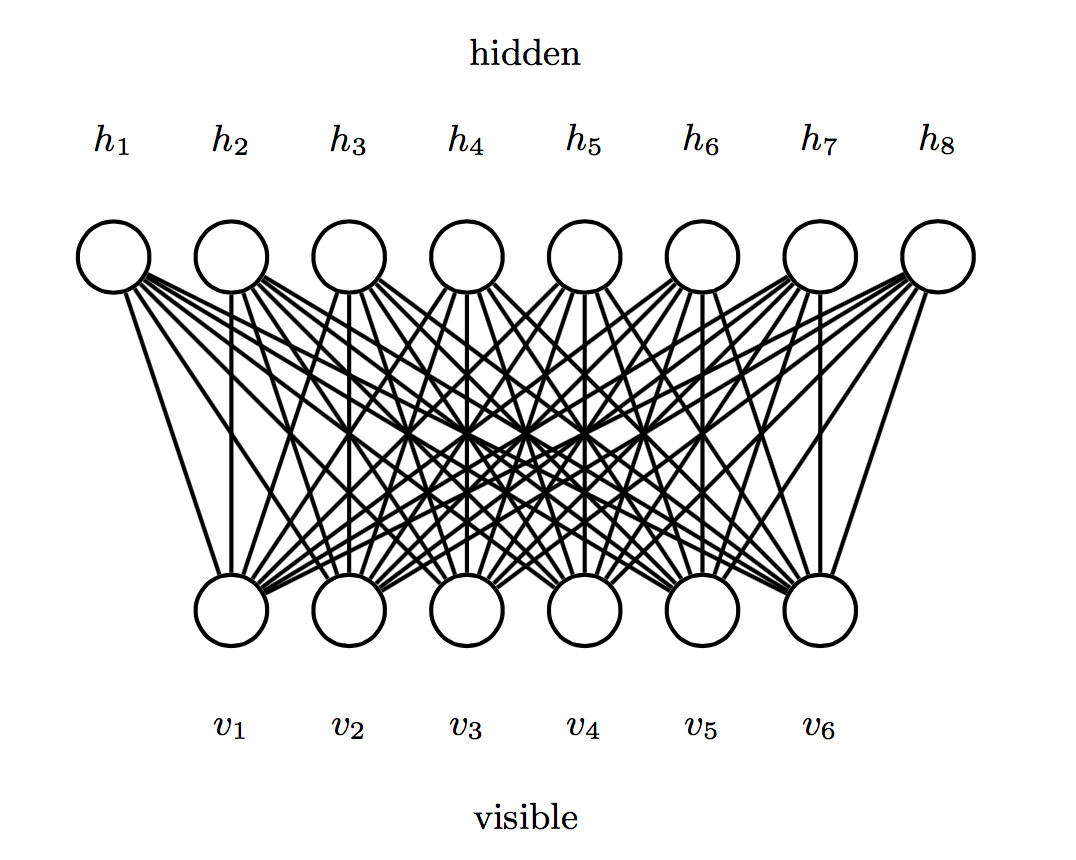}
   \caption{\label{fig:example-unsup} Example of unsupervised learning with a single RBM.}
\end{figure}

We now show an example for unsupervised learning. The network architecture is a single RBM with six visible units and eight hidden units; see Fig.~\ref{fig:example-unsup}. The goal is to learn a simple pattern (\verb|Pattern|) as shown below within the script in Fig.~\ref{fig:example-unsup-script}. This \verb|Pattern| is a very simple example of unlabeled data.

All the computational times reported throughout this section were obtained running on a personal computer with the following characteristics: 2.3 GHz Intel Core i7 and 4GB memory; Linux-Ubuntu~12.04 and GNU Octave~3.2.4. 

\subsubsection{Script}
Fig.~\ref{fig:example-unsup-script} shows the Matlab/Octave script for our example of unsupervised learning. We used the same toolbox than for the MNIST database \cite{Tanaka:2014:RBM-toolbox-Matlab}.  
\begin{figure}
\centering
\begin{verbatim}
% Matlab/Octave script
  clear all;
  addpath('..');
  ini = clock;
  Pattern = [1   1   0   0   0   0 ;
             0   0   1   1   0   0 ;
             0   0   0   0   1   1 ];                     
  Inputs = 6;       % #no variables as input
  TrainData = Pattern(:,1:Inputs)
  TestData = TrainData;
  nodes = [Inputs 8]; % [#inputs #hidden]
  bbdbn = randDBN( nodes, 'BBDBN' ); % Bernoulli-Bernoulli RBMs
  nrbm = numel(bbdbn.rbm);
  % meta-paramters or hyper-parameters
  opts.MaxIter = 50;
  opts.BatchSize = 1;
  opts.Verbose = false; 
  opts.StepRatio = 2.5;
  opts.object = 'CrossEntorpy'; 
  %Learning stage
  fprintf( 'Training...\n' );
  opts.Layer = nrbm-1;
  bbdbn = pretrainDBN(bbdbn, TrainData, opts);
  %Testing stage
  fprintf( 'Testing...\n' );
  H = v2h( bbdbn, TestData); % visible layer to hidden layer
  out = h2v(bbdbn,H); % hidden to visible layers (reconstruction)
  fprintf( 'Results...\n' );
  out
  round(out)
  theend = clock;
  fprintf('\nElapsed time: %f\n',etime(theend,ini));
\end{verbatim}
\caption{Matlab/Octave script: unsupervised learning example.}
\label{fig:example-unsup-script}
\end{figure}
\subsubsection{Results}
As the script in Fig.~\ref{fig:example-unsup-script} is executed, several data is displayed.
First, the training data (our pattern) is shown:
\begin{Output}[H]
\begin{verbatim}
TrainData =
   			1   1   0   0   0   0
   			0   0   1   1   0   0
   			0   0   0   0   1   1
\end{verbatim}  
\end{Output}
then the output after \verb|pretrainDBN| is: 
\begin{Output}[H]
\begin{verbatim}
out =
   		9.9e-01   9.9e-01   1.0e-04   1.2e-04   2.5e-05   2.6e-05
   		5.0e-04   4.3e-04   9.9e-01   9.9e-01   6.1e-04   6.0e-04
   		4.2e-06   3.6e-06   5.9e-06   5.4e-06   9.9e-01   9.9e-01
\end{verbatim}
\end{Output}
This output are the probabilities [0,1], known as reconstructions, so we apply the function \verb|round|, and then we obtain:
\begin{Output}[H]
\begin{verbatim}
out =
   		1   1   0   0   0   0
   		0   0   1   1   0   0
   		0   0   0   0   1   1
\end{verbatim}
\end{Output}
The total elapsed time is 0.254730 seconds.

\subsubsection{Discussion}
We found that the DNN in Fig.~\ref{fig:example-unsup} is able to learn the given pattern in an unsupervised manner. We use a \verb|stepRatio| of 2.5 because this allow us to have fewer iterations, i.e.\ \verb|MaxIter = 50|. Some authors recommend a learning rate (\texttt{stepRatio}) of 0.01 \cite{Hinton:2010:Guide,Tanaka:2014:RBM-toolbox-Matlab}, but with this setting, we need at least 1,000 iterations to learn the pattern; see \S\ref{sec:examples:XOR} for parameter setting issues.

\subsection{Predicting Patterns}\label{sec:examples:patterns}
\begin{figure}[H]
\centering
\includegraphics[width=4in]{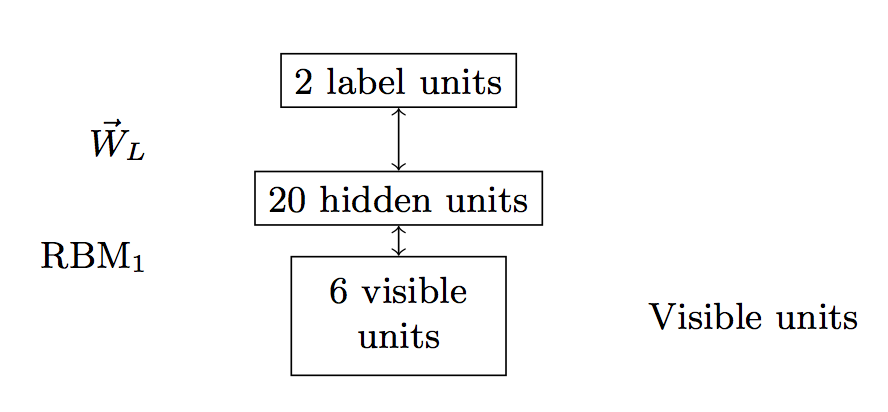}
\caption{\label{fig:example-patterns} A DNN architecture for the example of supervised learning: predicting patterns.}
\end{figure}

The following example simulates a time series prediction scenario. We test two different patterns (\texttt{Pattern1} and \texttt{Pattern2}). Our training data is a matrix with eight columns, which is the number of variables. We use only six variables as input, and the last two column variables as output. The main idea is that we feed the network only with six variables, then the network must ``predict'' the next two variables; see Fig.~\ref{fig:example-patterns}. Compared with the previous example, we use supervised learning as shown above in the MNIST example, see \S\ref{sec:toolbox}. Therefore, the labels are our two last columns of the pattern (outputs), i.e.~\verb|TrainLabels|.

\subsubsection{Script}
The script for this example is shown in Fig.~\ref{fig:example-patterns}.
\begin{figure}
\centering
\begin{verbatim}
% Matlab/Octave script
  clear all;
  addpath('..');
  ini = clock();
  Pattern1 = [ 0 0 0 0 0 1 0 1;
               0 0 0 0 1 0 1 0; 
               0 0 0 1 0 1 0 0; 
               0 0 1 0 1 0 0 0;
               0 1 0 1 0 0 0 0;
               1 0 1 0 0 0 0 0];
               
  Pattern2 = [ 1 1 0 0 0 0 0 0;
               0 1 1 0 0 0 0 0; 
               0 0 1 1 0 0 0 0; 
               0 0 0 1 1 0 0 0;
               0 0 0 0 1 1 0 0;
               0 0 0 0 0 1 1 0;
               0 0 0 0 0 0 1 1;
               1 0 0 0 0 0 0 1]; 
               
  Pattern=Pattern1;  % set the pattern to simulate                
  Inputs = 6;    % #no variables as input
  Outputs = 2;   % #no. variables as ouputs                        
  TrainData    = Pattern(:,1:Inputs);  % get the training data, inputs
  TrainLabels = Pattern(:,Inputs+1:Inputs+Outputs); % show the labels
  TestData = TrainData;      % we test with the same training data
  TestLabels = TrainLabels;  
  nodes = [Inputs 20 Outputs]; % [#inputs #hidden #outputs]
  bbdbn = randDBN( nodes, 'BBDBN' ); % Bernoulli-Bernoulli RBMs
  nrbm = numel(bbdbn.rbm);
  % meta-parameters or hyper-parameters
  opts.MaxIter = 1000;
  opts.BatchSize = 6;
  opts.Verbose = false;  
  opts.StepRatio = 2.5;
  opts.object = 'CrossEntorpy'; 
  %Learning stage
  fprintf( 'Training...\n' );
  [TrainData TrainLabels]
  opts.Layer = nrbm-1;
  bbdbn = pretrainDBN(bbdbn, TrainData, opts);
  bbdbn= SetLinearMapping(bbdbn, TrainData, TrainLabels);
  opts.Layer = 0;
  bbdbn = trainDBN(bbdbn, TrainData, TrainLabels, opts);
  fprintf( 'Testing...\n' );
  out = v2h( bbdbn, TestData );
  [TestData out]
  [TestData round(out)]
  theend = clock();
  fprintf('\nElapsed time: %f\n',etime(theend,ini));
\end{verbatim}
\caption{Matlab/Octave script: predicting patterns example.}
\label{fig:example-patterns-script}
\end{figure}

\subsubsection{Results}
The script in Fig.~\ref{fig:example-patterns-script} generates the following output. First, it prints out the \verb|TrainData| and \verb|TrainLabels| together with the instruction \verb|[TrainData TrainLabels]|:
\begin{Output}[H]
\begin{verbatim}
Training...
ans =
   		0   0   0   0   0   1   0   1
   		0   0   0   0   1   0   1   0
   		0   0   0   1   0   1   0   0
   		0   0   1   0   1   0   0   0
   		0   1   0   1   0   0   0   0
   		1   0   1   0   0   0   0   0
\end{verbatim}
\end{Output}
The reconstructions (probabilities) are given in the last two columns:
\begin{Output}[H]
\begin{verbatim}
Testing...
ans =
		0.0000 0.0000 0.0000 0.0000 0.0000 1.0000 0.0031 0.9881
		0.0000 0.0000 0.0000 0.0000 1.0000 0.0000 0.9911 0.0011
		0.0000 0.0000 0.0000 1.0000 0.0000 1.0000 0.0044 0.0112
		0.0000 0.0000 1.0000 0.0000 1.0000 0.0000 0.0092 0.0073
		0.0000 1.0000 0.0000 1.0000 0.0000 0.0000 0.0065 0.0002
		1.0000 0.0000 1.0000 0.0000 0.0000 0.0000 0.0003 0.0041
\end{verbatim}
\end{Output}
As in the previous example, we apply the function \verb|round|, so we obtain:
\begin{Output}[H]
\begin{verbatim}
[TestData round(out)]
ans =
   		0   0   0   0   0   1   0   1
   		0   0   0   0   1   0   1   0
   		0   0   0   1   0   1   0   0
   		0   0   1   0   1   0   0   0
   		0   1   0   1   0   0   0   0
   		1   0   1   0   0   0   0   0
\end{verbatim}
\end{Output}
\subsubsection{Discussion}
In this example, we set a DNN to predict two variables given six input variables. We found that the DNN is able to successfully predict the given patterns. Here, we show only results for \texttt{Pattern1}, but the results for \texttt{Pattern2} are similar. We started to test the DNN with a different pattern, and we came across accidentally with a pattern that the DNN cannot predict (4 inputs, 2 outputs):
\begin{Output}[H]
\begin{verbatim}
Training...
ans =
   		0   0   0   0   0   1
   		0   0   0   0   1   0
   		0   0   0   1   0   0
   		0   0   1   0   0   0
   		0   1   0   0   0   0
   		1   0   0   0   0   0
\end{verbatim}
\end{Output}        
\begin{Output}[H]
\begin{verbatim}        
Testing...
ans =
   		0.00000   0.00000   0.00000   0.00000   0.49922   0.49922
   		0.00000   0.00000   0.00000   0.00000   0.49922   0.49922
   		0.00000   0.00000   0.00000   1.00000   0.00993   0.00993
   		0.00000   0.00000   1.00000   0.00000   0.01070   0.01070
   		0.00000   1.00000   0.00000   0.00000   0.01142   0.01142
   		1.00000   0.00000   0.00000   0.00000   0.01109   0.01109
\end{verbatim}
\end{Output}  

The two first rows of the training data have the same input values, but they have different output. Therefore, the DNN ``intelligently'' suggest an output of 0.499 (probability).

\subsubsection{XOR problem}\label{sec:examples:XOR}
The XOR problem is a non-linear problem that is typical test for a classifier because it is a problem that a simple linear classifier cannot learn. In the neural networks literature, an example of a linear classifier is the \emph{perceptron} introduced by Frank Rosenblatt in 1957~\cite{Rosenblatt:1957:perceptron}. A decade later, Marvin Minsky and Seymour Paper wrote their famous book \emph{Perceptrons}, and they showed that perceptrons cannot solve the XOR problem \cite{Minsky:1969:book}. Perhaps partly due to the publication of \emph{Perceptrons} , there was a decline of research in neural networks until the backpropagation algorithm appeared about twenty year after Minsky and Paper's publication. 

\begin{figure}
\centering
\includegraphics[width=4in]{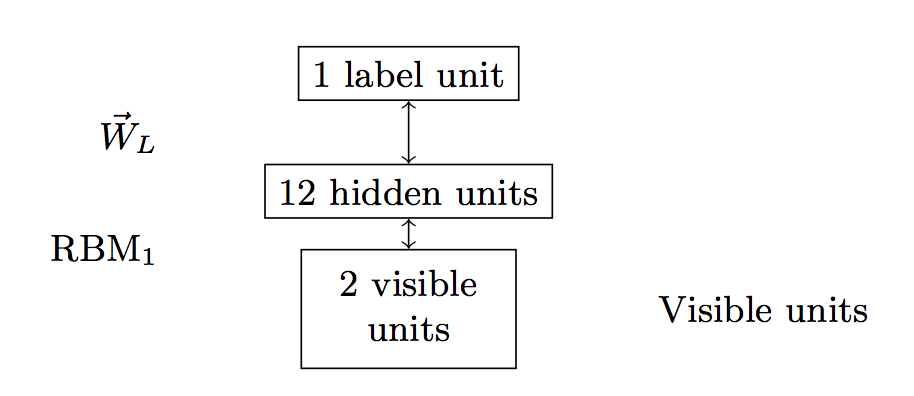}
\caption{\label{fig:example-XOR} A DNN architecture for the XOR example.}
\end{figure}

Here, we analyze the XOR problem with a DNN; see Fig.~\ref{fig:example-XOR}. 

\subsubsection{Script}
The script for the XOR problem is shown in Fig.~\ref{fig:example-XOR-script}.
\begin{figure}[H]
\centering
\begin{verbatim}
%Matlab/Octave script
  TrainData    = [0 0; 0 1; 1 0; 1 1];
  TrainLabels = [0; 1; 1; 0];
  TestData = TrainData;
  TestLabels = TrainLabels; 
  nodes = [2 12 1]; % [#inputs #hidden #outputs]
  bbdbn = randDBN( nodes, 'BBDBN' ); %  Bernoulli-Bernoulli RBMs    
  nrbm = numel(bbdbn.rbm);
  % meta-paramters or hyper-parameters
  opts.MaxIter = 100;
  opts.BatchSize = 4;
  opts.Verbose = false;  
  opts.StepRatio = 2.5;
  opts.object = 'CrossEntorpy'; 
  %Learning stage
  fprintf( 'Training...\n' );
  opts.Layer = nrbm-1;
  bbdbn = pretrainDBN(bbdbn, TrainData, opts);
  bbdbn= SetLinearMapping(bbdbn, TrainData, TrainLabels);
  opts.Layer = 0;
  bbdbn = trainDBN(bbdbn, TrainData, TrainLabels, opts);
  %Testing stage
  fprintf( 'Testing...\n' );
  TestData =  [0.1 0.1; 0 0.9; 1 0.2; 0.8 1];
  TestData
  out = v2h( bbdbn, TestData )
  %Printing results
  fprintf( '\nResults:\n\n' );
  rmse= CalcRmse(bbdbn, TestData, TestLabels);
  ErrorRate= CalcErrorRate(bbdbn, TestData, TestLabels);
  fprintf( 'For test data:\n' );
  fprintf( 'rmse: %g\n', rmse );
  fprintf( 'ErrorRate: %g\n', ErrorRate );
\end{verbatim}
\caption{Matlab/Octave script: XOR example.}
\label{fig:example-XOR-script}
\end{figure}

\subsubsection{Results}
The script in Fig.~\ref{fig:example-XOR-script} involves only two input variables and a single output, so the \verb|TrainData| and \verb|TrainLabels| are as follows:
\begin{Output}[H]
\begin{verbatim}
TrainData =
   			0   0   
   			0   1  
   			1   0
   			1   1
\end{verbatim}
\end{Output}
\begin{Output}[H]
\begin{verbatim}
TrainLabels =
   				0   
   				1   
   				1   
   				0
\end{verbatim}
\end{Output}

Before coming up with the script of Fig.~\ref{fig:example-XOR-script}, we tested different configurations for the DNN. We started with the following hyperparameter setting:
\begin{Output}[H]
\begin{verbatim}
	nodes = [2 3 3 1]; % [#inputs #hidden #hidden #outputs]
	pts.MaxIter = 10;
	opts.BatchSize = 4;
	opts.Verbose = true;
	opts.StepRatio = 0.1;
	opts.object = 'CrossEntropy';
	TestData = TrainData;
\end{verbatim}
\end{Output}
The \verb|TrainLabels| is a column vector, and \verb|out'| is a row vector, where \verb|out'| is the transpose of \verb|out|. Therefore, the desired output is \verb|out' =  0 1 1 0|. The output of the DNN for this setting is:
\begin{verbatim}
       out’ = 0.49994   0.49991   0.50009   0.50006
\end{verbatim}
This setting does not work with the above parameters. If we add more iterations \verb|opts.MaxIter = 100|, it still does not work properly, and we obtain the output:
\begin{verbatim}
       out’ = 0.47035   0.51976   0.49161   0.50654
\end{verbatim}
If we add more hidden neurons \verb|nodes = [2 12 12 1]| with the same number of iterations, then the performance improves. The output now is:
\begin{verbatim}
       out’ = 0.118510   0.906046   0.878771   0.096262
\end{verbatim}
The performance is still better if we add more iterations \verb|opts.MaxIter = 1000|. The output now is:
\begin{verbatim}
       out’ = 0.014325   0.982409   0.990972   0.012630
       Elapsed time: 32.607048 seconds
\end{verbatim}
The previous experiment takes about 33 seconds. In order to decrease the complexity of the network, we reduce the number of hidden neurons. Now we have a single hidden layer \verb|nodes = [2 12 1]|, and with the same number of iterations \verb|opts.MaxIter = 1000|. In about 24 seconds, the output is:
\begin{verbatim}
       out’ = 0.043396   0.950205   0.947391   0.059305
       Elapsed time: 23.640984 seconds
\end{verbatim}

Now, if we reduce the the number of iterations with less neurons as the previous experiments, i.e.\ 
\verb|opts.MaxIter = 100|, it is faster but the performance decays. So the output now is:
\begin{verbatim}
       out’ =  0.16363   0.80535   0.82647   0.20440
       Elapsed time: 2.617439 seconds
\end{verbatim}

Other hyperparameter is \verb|opts.StepRatio|, the learning rate, so we tunned this parameter to \verb|opts.StepRatio = 0.01|, and the number of iterations is \verb|opts.MaxIter = 1000|. We found similar results to the previous experiment, i.e.\ we do not reach the desired output. 

Nevertheless, if we use \verb|opts.StepRatio = 2.5| and \verb|opts.MaxIter = 100|, then we obtain a good performance in about one second. The output is:
\begin{verbatim}
       out’ = 0.022711   0.955236   0.955606   0.065202
       Elapsed time: 0.806343
\end{verbatim}

Another important experiment is to test the performance with real values. So we change the test data, and we obtain the following results:

\begin{Output}[H]
\begin{verbatim}
  TestData =
            0.10000   0.10000
            0.00000   0.90000
            1.00000   0.20000
            0.80000   1.00000
\end{verbatim}
\end{Output}            
\begin{Output}[H]
\begin{verbatim}
  out =
            0.213813
            0.956192
            0.889679
            0.053432

  Elapsed time: 0.686760
\end{verbatim}
\end{Output}

Finally, we ran some experiments by tuning the hyperparameter \verb|opts.object| to `Square'  or  `CrossEntorpy', and we found no difference in performance. The flag \verb|opts.verbose| is only for showing or not the training performance.

\subsubsection{Discussion}
For the XOR problem, we found that the best performance is with the following combination: \verb|opts.StepRatio = 2.5| and \verb|opts.MaxIter = 100|. A large step ratio with few iterations allow us to obtain results faster than other settings. The performance is good enough to solve the XOR problem. Moreover, this setting classifies correctly when the inputs are real data. For this reason, this setting is used in the previous examples; see \S\ref{sec:examples:unsupervised} and \S\ref{sec:examples:patterns}.

\section{Speech Processing}\label{sect:speech}
\begin{figure*}
\centering
\includegraphics[width=4.5in]{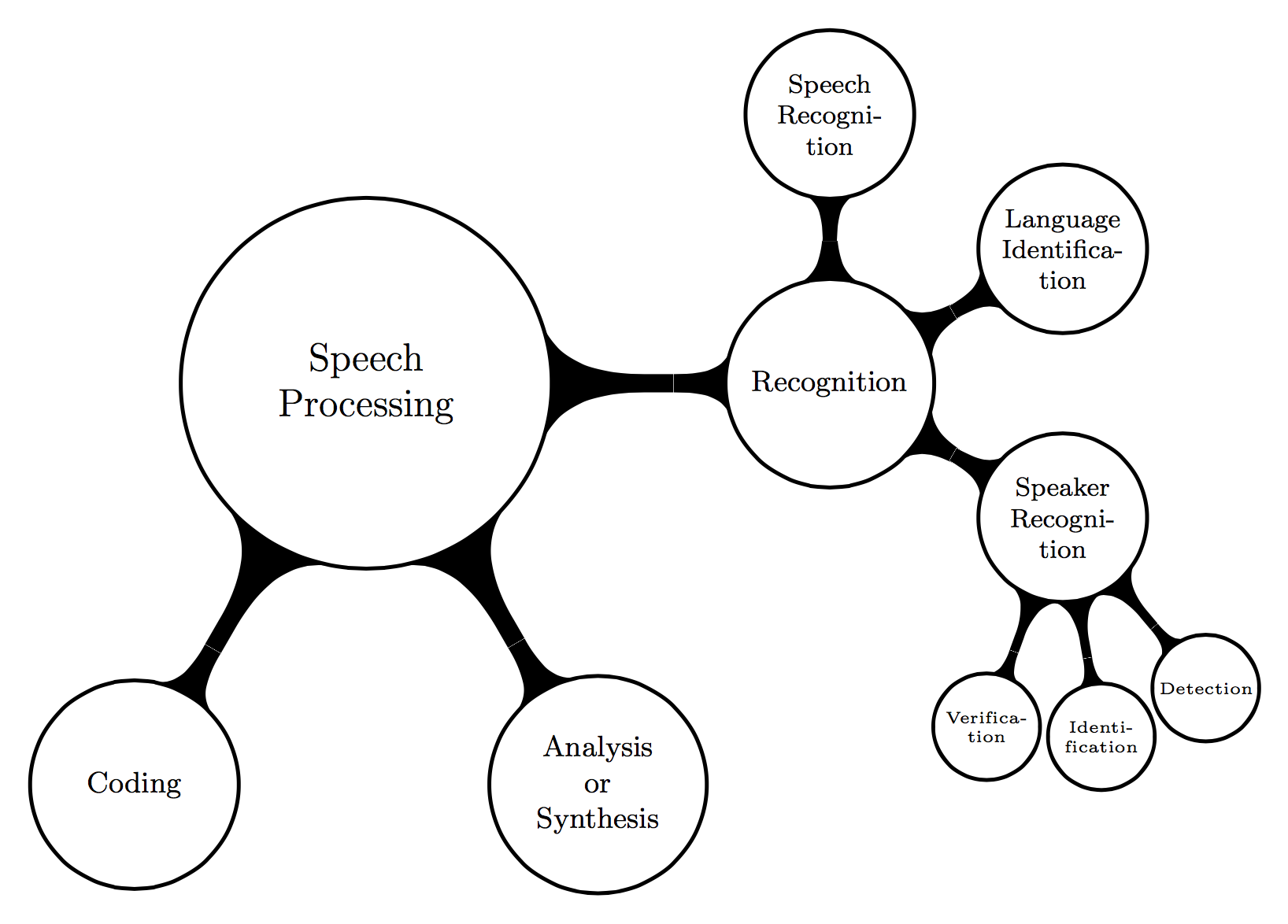}
\caption{\label{fig:speech}Speech Processing}
\end{figure*}

Speech Processing has several applications including Speech Recognition, Language Identification and Speaker Recognition; see Fig.~\ref{fig:speech}. Sometimes, additional information is stored and associated to speech. Therefore, Speaker Recognition can be either text dependent or text independent. Moreover, Speaker Recognition involves different tasks such as \cite{Campbell:1997:IEEE}:
\begin{itemize}
\item Speaker Identification 
\item Speaker Detection
\item Speaker Verification. 
\end{itemize}

\subsection{Speech Features}
\begin{figure*}
\centering
\includegraphics[width=4in]{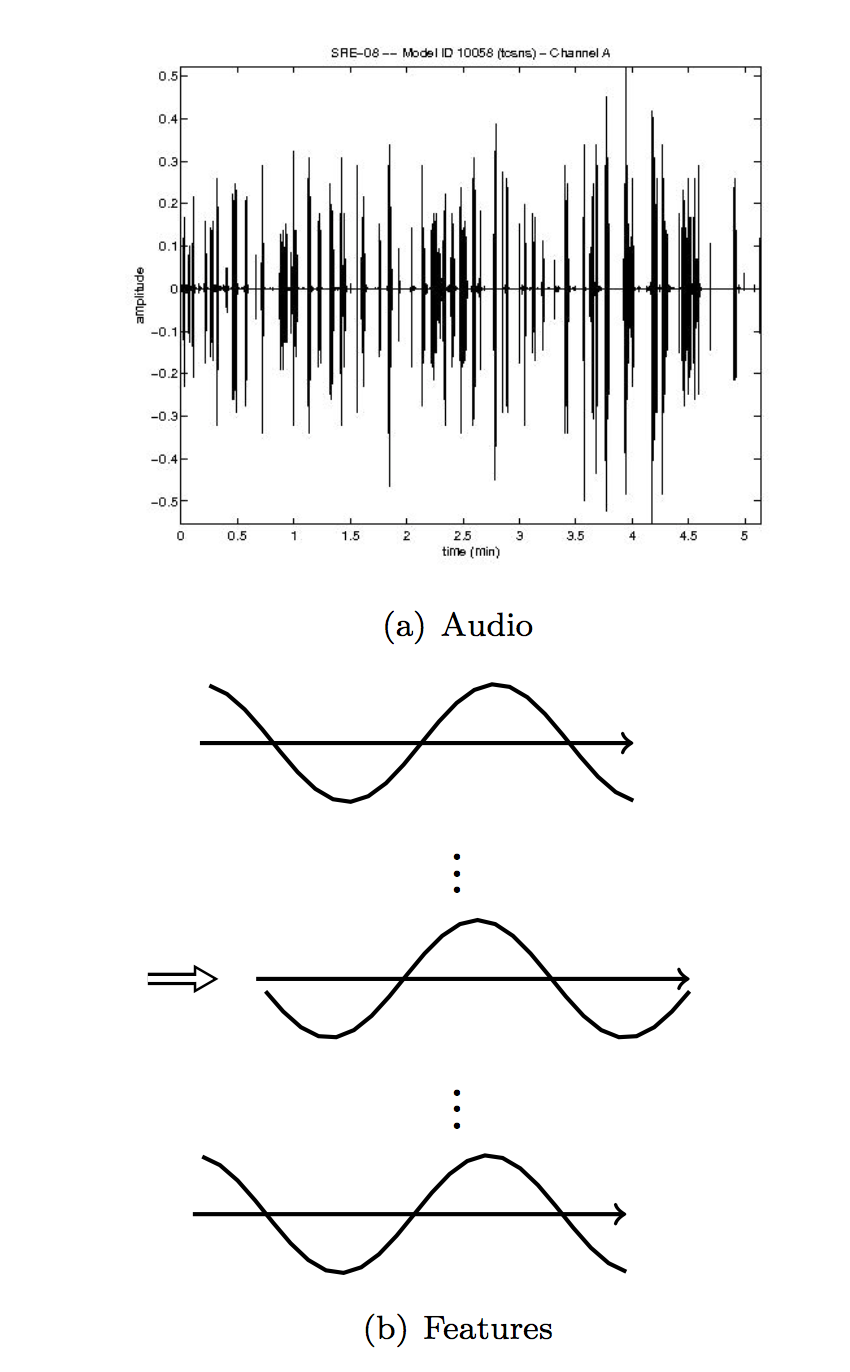}
\caption{Audio sample. (a) Audio represented as a time-domain waveform; (b) Features obtained from the time-domain waveform.}
\label{fig:audio-mfcc}
\end{figure*}

The first step for most speech recognition systems is the feature extraction from the time-domain sampled acoustic waveform (audio); see Figure \ref{fig:audio-mfcc}a. The time-domain waveform is represented by overlapping \textit{frames}. Each frame is generated every 10ms with a duration of 25ms. Then, a feature is extracted for every frame. Several methods have been investigated for feature extraction (acoustic representation) including Linear Prediction Coefficients (LPCs), Perceptual Linear Prediction (PLP) coefficients and Mel-Frequency spaced Cepstral Coefficients(MFCCs) \cite{Davis:1980:MFCCs,Togneri:2011:IEEE}.

\subsection{DNN and Speech Processing}
\begin{figure}
\centering
\includegraphics[width=3in]{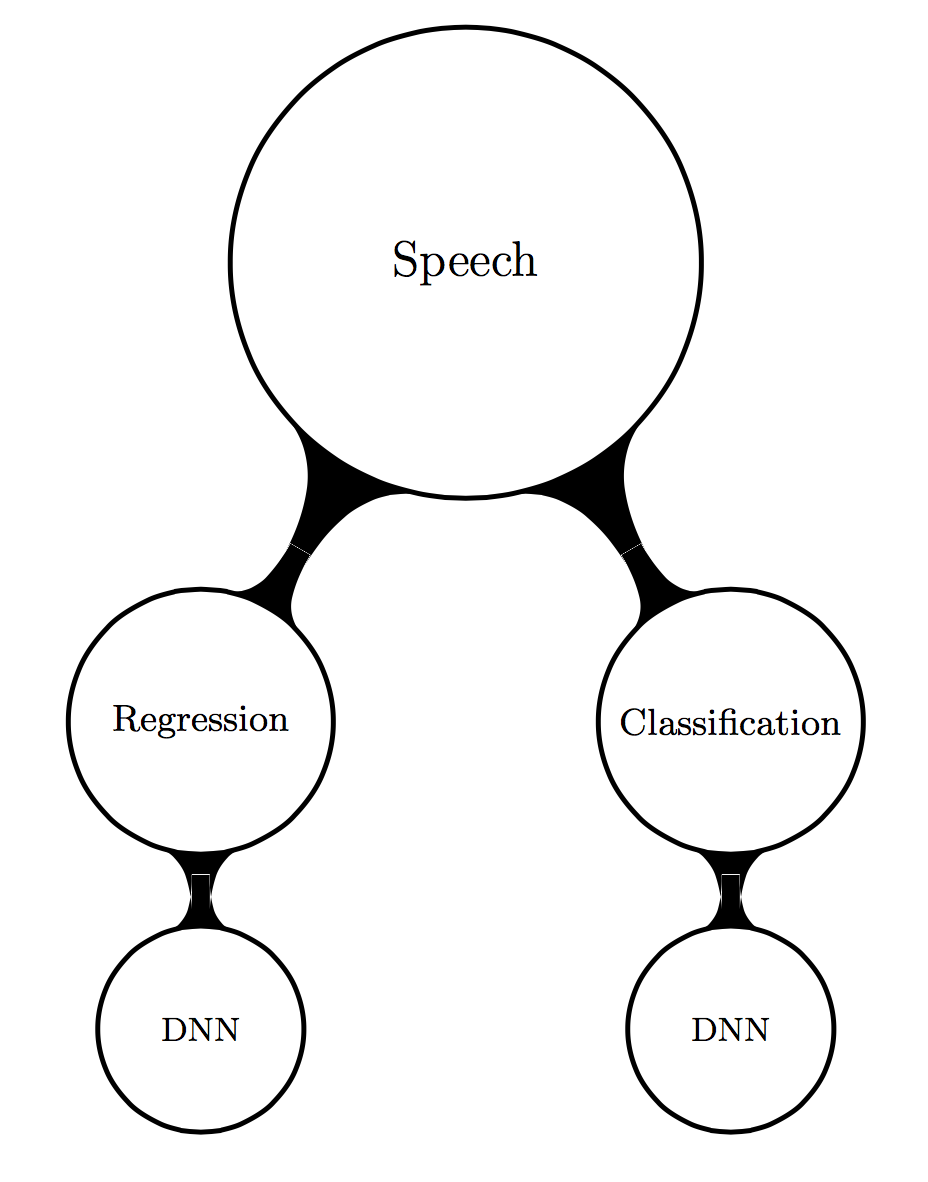}
\caption{\label{fig:DNN-speech}DNN and Speech}
\end{figure}

As we shown above, DNNs have the flexibility to be used as either unsupervised or supervised learning. Therefore, DNNs can be used for regression or classification problems in speech recognition; see Fig.~\ref{fig:DNN-speech}.

Nowadays, DNNs have been applied successfully on speech processing including speaker recognition \cite{Hinton:2012:DNN,Seltzer:2014:DNN-ICASSP}, language identification \cite{Gonzalez-Dominguez:2015:DNN} and speech generation \cite{Ling:2015:DNN-speech}.

VOICEBOX\footnote{http://www.ee.ic.ac.uk/hp/staff/dmb/voicebox/voicebox.html}
 is a Speech Processing Toolbox for MATLAB\textsuperscript{\textregistered{}}, which is also publicly available.

\section{Summary}\label{sec:summary}
Neural networks approaches have been used widely to build Intelligent Systems. We introduced Deep Neural Networks (DNNs) and Restricted Boltzmann Machines (RBMs), and their relationship to Deep Learning (DL) and Deep Belief Nets (DBNs). Across the literature, there are some introductory papers for RBMs \cite{Hinton:2010:Guide,Fischer:2014:RBM}. One of the contributions of this tutorial are the simple examples for a better understanding of RBMs and DNNs. The examples cover unsupervised and supervised learning, therefore, we cover both unlabeled and labeled data, for prediction and classification. Moreover, we introduce a publicly available MATLAB\textsuperscript{\textregistered{}} toolbox to show the performance of DNNs and RBMs \cite{Tanaka:2014:RBM-toolbox-Matlab}. The toolbox and the examples have been tested on Octave, the open source version of MATLAB\textsuperscript{\textregistered{}}. The last example, XOR problem, presents some results by different setting of some hyperparameters of DNNs. Finally, two applications for intelligent pattern recognition are also covered on this tutorial: the MNIST benchmarking and speech recognition.

\bibliography{biblio}
\bibliographystyle{plain}
\end{document}